\PassOptionsToPackage{table}{xcolor}

\documentclass[a4paper, 10pt, conference]{ieeeconf}      
\usepackage{FG2024}
\usepackage{times}
\usepackage{epsfig}
\usepackage{graphicx}
\usepackage{amsmath}
\usepackage{amssymb}
\usepackage{multirow}
\usepackage{xcolor}
\usepackage{booktabs}
\usepackage{gensymb}
\definecolor{maroon}{cmyk}{0,0.87,0.68,0.32}

\usepackage[pagebackref,breaklinks,colorlinks]{hyperref}
\FGfinalcopy 

\IEEEoverridecommandlockouts                              
\def\FGPaperID{****} 

\title{\LARGE \bf
    The Paradox of Motion: Evidence for Spurious Correlations in Skeleton-based Gait Recognition Models
}

\author{\parbox{16cm}{\centering
   {\large Andy Cătrună, Adrian Cosma, Emilian Rădoi}\\
    {\tt\small andy\_eduard.catruna@upb.ro, ioan\_adrian.cosma@upb.ro, emilian.radoi@upb.ro}\\
   {\normalsize
   Faculty of Automatic Control and Computer Science, POLITEHNICA Bucharest}}
}

\begin{document}

\ifFGfinal
\thispagestyle{empty}
\pagestyle{empty}
\else
\author{Anonymous FG2024 submission\\ Paper ID \FGPaperID \\}
\pagestyle{plain}
\fi
\maketitle

\begin{abstract}
Gait, an unobtrusive biometric, is valued for its capability to identify individuals at a distance, across external outfits and environmental conditions. This study challenges the prevailing assumption that vision-based gait recognition, in particular skeleton-based gait recognition, relies primarily on motion patterns, revealing a significant role of the implicit anthropometric information encoded in the walking sequence. We show through a comparative analysis that removing height information leads to notable performance degradation across three models and two benchmarks (CASIA-B and GREW).
Furthermore, we propose a spatial transformer model processing individual poses, disregarding any temporal information, which achieves unreasonably good accuracy, emphasizing the bias towards appearance information and indicating spurious correlations in existing benchmarks. These findings underscore the need for a nuanced understanding of the interplay between motion and appearance in vision-based gait recognition, prompting a reevaluation of the methodological assumptions in this field. Our experiments indicate that "in-the-wild" datasets are less prone to spurious correlations, prompting the need for more diverse and large scale datasets for advancing the field.
\end{abstract}

\section{Introduction}
\label{sec:intro}
Gait recognition, the method of identifying people based on their unique walking patterns, has become a significant area of research due to its benefits compared to other privacy-intrusive approaches \cite{sepas2022deep}. These benefits amount to not requiring subject cooperation, utilizing data that is more privacy-friendly when compared to the faces, iris, or fingerprints, and good performance at a distance.

In general, gait analysis is an interdisciplinary field that combines elements of biomechanics, kinesiology, medicine, and engineering to study and evaluate human walking patterns \cite{nixon2005gait}. In academic research, gait analysis involves the systematic examination of various aspects of human locomotion to gain insights into normal and pathological walking patterns as well as individual patterns related to identity and demographic information such as gender and age. Gait analysis primarily focuses on the study of \textit{movement patterns across time} rather than appearance information or body shape. The goal is to understand the biomechanics of walking and the dynamic aspects of human locomotion. The focus on movement patterns distinguishes gait analysis from disciplines that primarily deal with body shape or appearance, such as anthropometry or the study of body measurements such as height, weight and limb proportions. 

\begin{figure}[hbt!]
    \centering
    \includegraphics[width=0.75\linewidth]{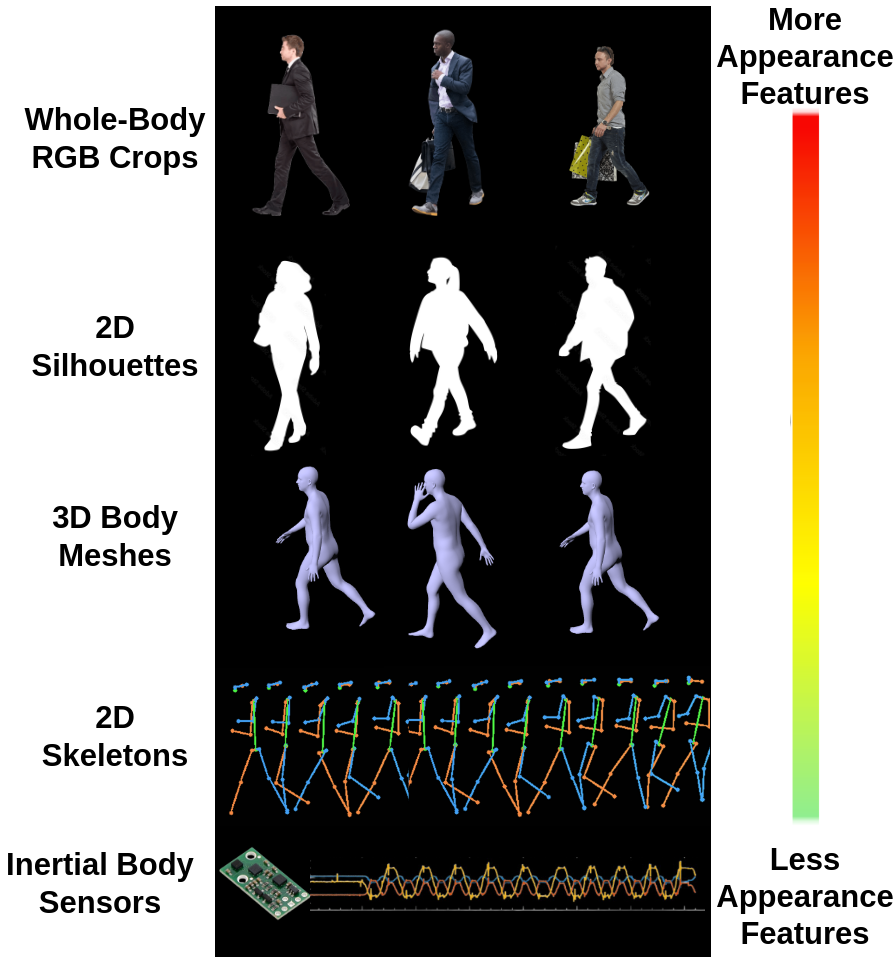}
    \caption{Examples of common modalities used for person recognition in biometrics literature ordered by the amount of appearance information present in each one. Intuitively, sequences of 2D skeletons extracted with pose estimation models contain the least amount of appearance features. However, our experiments show that body proportions play a significant role in automatic recognition.}
    \label{fig:gait-appearance}
\end{figure}

In Figure \ref{fig:gait-appearance} we showcase the various modalities used in vision-based gait analysis for person recognition and re-identification across the field, ordered by amount of appearance information. Whole-body crops, in which the full person appearance is visible, are used in person re-identification \cite{ye2021deep}, clothes-changing person re-identification \cite{gu2022clothes} and, in some cases, gait-based recognition \cite{zhang2020learning}. Sequences of silhouettes are widely used in gait recognition with good performance, and they encode body shape, clothing and carrying conditions alongside hairstyle and other details. Body meshes such as SMPL \cite{smpl} abstract away clothing and carrying conditions but, by their construction contains body shape and body composition information. Body meshes have been used in gait-recognition in more recent methods \cite{li2020end,zheng2022gait,cosma2023psymo}. Skeletons extracted from pose estimation models \cite{xu2022vitpose} are constructed to be body-shape agnostic, ignoring overt clothing, carrying conditions and body composition, but do encode height and body proportions of the subject. Lastly, inertial body sensors encode only movement information across time, but are unfeasible to be used in the real recognition scenarios as they are obtrusive and require extensive subject cooperation. Consequently, skeletons offer the best trade-off between usage feasibility in real world, unobtrusiveness and privacy. However, in this paper, we show that crucial and discriminative appearance information is stored in skeletons. This data aids person recognition beyond motion patterns, raising concerns about privacy and questioning the current progress in vision-based gait recognition literature. 

As the recognition models have access to both movement information and appearance data, we cannot exclude the possibility of them focusing only on appearance features and discarding the gait. There are hints of this behaviour in multiple works on appearance-based gait recognition. For instance, GaitSet \cite{chao2019gaitset} treats the sequence of silhouettes as an unordered set. Their results show that impressive recognition accuracy can be obtained with as few as 5 consecutive frames which do not contain enough gait information. Similarly, Xu et al. \cite{xu2020gait} demonstrate good performance for the gender estimation using only one silhouette. Moreover, the sub-field of clothes-changing person re-identification (CC-ReID) \cite{gu2022clothes} utilize a single crop of a person to perform recognition across multiple outfits by extracting clothes-irrelevant features. The promising performance of these approaches hint that performance for gait recognition models (in some scenarios, reaching 99\% accuracy \cite{zou2023multi} on benchmarks such as CASIA-B \cite{CASIA:Yu}) is not solely attributed to movement patterns, but rather to anthropometric features such as height, body proportions and composition. This effect is attributed to spurious correlations and appearance information leakage from modality extractors.

In this work, we shed light on two research questions: \textbf{RQ1:} Do indirect anthropometric features such as height and body proportions play a role in the promising performance of gait recognition models? \textbf{RQ2:} Are there spurious correlations in current gait recognition benchmarks that inflate reported results? 

In the interest of answering \textbf{RQ1}, we present a comparative study of the normalization methods for skeleton sequences utilized by state-of-the-art methods as well as normalization methods inspired from image processing tasks. Our results on the popular gait recognition benchmarks CASIA-B \cite{CASIA:Yu} and GREW \cite{GREW:zhu} show that normalization techniques that remove height and location information significantly reduce the performance. 

Furthermore, for answering \textbf{RQ2}, we construct a model that only utilizes a single pose for the prediction without having any motion information. Surprisingly, our results show that a carefully constructed model that only processes poses individually obtains a significant recognition accuracy of 57.49\% on CASIA-B \cite{CASIA:Yu} and a Rank-1 accuracy of 28.19\% on GREW \cite{GREW:zhu} by following the exact evaluation protocol as for gait recognition. This proves that skeleton-based gait recognition methods do not focus primarily on the gait information and are heavily reliant on appearance information leaked by upstream modality extractors. 

This work makes the following contributions:

\begin{enumerate}
    \item We show that skeletons extracted from pose estimation models encode appearance features that aid recognition beyond motion patterns. Our experiments indicate that removing height information through different sequence scaling procedures severely impacts gait recognition performance on two popular benchmarks \cite{CASIA:Yu,GREW:zhu} across three models \cite{teepe2021gaitgraph,cosma2022learning,catruna2023gaitpt}.
    \item We further provide evidence for this behaviour by training a spatial transformer model on individual skeletons, completely disregarding temporal information. We obtain unreasonably high performance only through exploiting spatial relationships between joints.
    \item Our experiments indicate that CASIA-B, the most popular benchmark for vision-based gait recognition on which current methods have saturated performance, exhibits spurious correlations. This calls into question the current progress in gait recognition in controlled settings. Simply removing the sequence position in the frame markedly lowers performance
\end{enumerate}

\section{Related Work}
\label{sec:related}
The approaches for gait recognition are typically divided into two categories: appearance-based methods \cite{chao2019gaitset,xu2020gait,zheng2022gait,lin2021gait} and model-based methods \cite{liao2020model,liao2017pose,cosma2022learning,teepe2021gaitgraph,cyclegait2023,li2020jointsgait,gpgait2023}. Appearance-based methods usually employ background subtraction or instance segmentation algorithms \cite{hafiz2020survey} to obtain the silhouettes of people and process them with tailored neural network architectures \cite{chao2019gaitset,xu2020gait}. These approaches are explicitly relying on appearance information, as they make use of body shape across the gait cycle. Silhouette-based methods typically do not have a strong enough mechanism of forcing the model to only analyse the gait and disregard the appearance, making the neural network to learn to mostly disregard temporal dynamics of gait and focus only on body shape for recognition. Consequently, silhouette-based method are part of a boarder category of person identification methods termed "whole body recognition" \cite{huang2023whole,myers2023recognizing,1240273}. In some adverse scenarios, such as gait identification at a large distance (e.g. on the BRIAR dataset \cite{huang2023whole}), it is difficult to properly isolate fine-grained movements \cite{huang2023whole}, and methods explicitly rely on body shape by, for example, fitting a 3D SMPL mesh \cite{thakkar2021feasibility,zheng2022gait} on the person.

Furthermore, an emerging area of research is clothes-changing person re-identification (CC-ReID) \cite{gu2022clothes,arkushin2022reface,jin2022cloth,yu2020cocas}, which aims to recognize an individual from a single image while having different outfits. However, in this scenario, it is clear that models must take into account body shape, height and body composition to construct outfit-invariant features. For instance, features maps from the method developed by Gu et al., \cite{gu2022clothes} show that models rely on facial attributes and overall body shape for a robust prediction.

Conversely, model-based methods obtain a human model from the sequence of images in a video, typically employing a human pose estimation model \cite{xu2022vitpose} which outputs a sequence of anatomical skeletons (i.e. containing 17 joint coordinates for the COCO format \cite{lin2015microsoft}). By discarding most appearance data except for the limb lengths, skeleton-based methods are more privacy-friendly than appearance-based methods and force the neural network to focus on the motion patterns. However, we posit that appearance features are still an important factor in the performance of skeleton-based gait recognition models, which resort to "shortcuts" by exploiting both anthropometric information and spurious correlations found in existing gait datasets \cite{CASIA:Yu,GREW:zhu}. In this work we show how simple scaling and translation of skeleton gait sequences severely impact performance, and that non-trivial recognition performance can be achieved using only a single pose.

In the area of skeleton-based gait recognition, there is no standard method for normalization of skeleton sequences. In Table \ref{tab:methods-norms} we showcase several popular methods and their respective normalization procedures. Methods employing scaling procedures that reduce height information such as PTSN \cite{liao2017pose}, PoseGait \cite{liao2020model} and GaitFormer \cite{cosma2022learning} have markedly lower performance than methods utilizing more neutral forms of normalization (i.e. learned batch normalization or normalization by frame coordinates) such as GaitGraph \cite{teepe2021gaitgraph}, CycleGait \cite{cyclegait2023} and JointsGait \cite{li2020jointsgait}. Perhaps unsurprisingly in our context, methods that include explicit anthropometric features (e.g., bone features) such as Gait-TR \cite{zhang2023spatial}, GaitGraph2 \cite{teepe2022towards} and GPGait \cite{gpgait2023} have good performance, regardless of type of scaling.

\begin{table}[hbt!]
    \centering
    \caption{Normalization procedures used in popular skeleton-based gait recognition models. Methods using batch normalization and frame scaling obtain better performance than skeleton level normalizations. The explicit addition of anthropometric features further boosts performance, regardless of normalization scheme.}
    \label{tab:methods-norms}
    \resizebox{0.85\linewidth}{!}{
    \begin{tabular}{l|p{1.5cm}p{2.2cm}}
        \textbf{Method} & \textbf{Normalization} & \textbf{Extra Features}\\
        \toprule
        PTSN \cite{liao2017pose} & Skel. Level  & - \\
        PoseGait \cite{liao2020model} & Skel. Level & -\\
        GaitFormer \cite{cosma2022learning} & Skel. Level & -\\
        \midrule
        GaitGraph \cite{teepe2021gaitgraph} &  Batch Norm & - \\
        CycleGait \cite{cyclegait2023} & Batch Norm & - \\
        JointsGait \cite{li2020jointsgait} & Batch Norm & -\\
        \midrule
        GaitPT \cite{catruna2023gaitpt} & Frame Scale / Seq. Level & -\\
        \midrule
        Gait-TR \cite{zhang2023spatial} & N/A & Motion velocities and  bone features \\
        GaitGraph2 \cite{teepe2022towards} &  Batch Norm. & Motion velocities and  bone features \\
        GPGait \cite{gpgait2023} & Skel. Level & Bone lengths and joint angles \\
    \end{tabular}
    }
\end{table}

In this work, we explore multiple scaling and translation methods for skeleton gait sequences and provide insights into the behaviour of popular skeleton-based gait recognition models. Furthermore, we show that a non-trivial gait recognition performance can be achieved with using only one skeleton per person, indicating the presence of spurious correlations in gait recognition datasets.

\section{Method}
\label{sec:method}
\subsection{Limiting Anthropometric Information through Gait Sequence Normalization}

We consider a gait sequence to be of the form $G=\{P_1, P_2,.., P_N\}$, $G \in R^{N \times V \times C}$ where $N$ is the number of frames in the sequence, $V$ is the number of joints in an individual pose, $C$ is the dimension of the joint coordinates, and $P_i = \{(x_1, y_1), (x_2, y_2),.., (x_{V}, y_{V})\}$, $P \in R^{V \times C}$ is a 2D pose consisting of $V$ joints, each joint being composed of a $x$ and $y$ coordinate. In our case, the number of frames in a sequence $N$ is equal to 60, the number of joints $V$ is 18 and the $C$ dimension is equal to 2. We analyse the effects of scaling-based and translation-based normalization techniques. We refer to scaling normalization as operations where the skeleton coordinates are divided by specific values so that the resulting coordinates fall within a defined range. Similarly, we utilize translation-based normalization techniques to center the coordinates around a predetermined value.

We analyse the following normalization techniques applied on the gait sequence:

\textbf{Normalizations that preserve height and position in frame:}
\begin{itemize}
    \item \textit{Frame Scale} - Scaling the $(x, y)$ coordinates of all skeletons with the same value - the width of the frame:
        \begin{equation}
            (x_i, y_i)_j = \frac{(x_i, y_i)_j}{f_W}
        \end{equation}
    where $f_W$ is a constant equal to the width of the video frame. $(x_i, y_i)_j$ corresponds to the $(x, y)$ coordinates of the $i$-th joint in the $j$-th pose in the sequence;
    \item \textit{Batch Norm} - Utilizing 2D Batch Normalization \cite{ioffe2015batch} for allowing the model to learn the scaling and translation coefficients for both the $X$ and $Y$ channel dimensions;
    \item \textit{Global Average Skeleton} - Translating the $(x, y)$ coordinates of all skeletons with the same value based on the average pelvis of the training set and scaling all skeletons with the same value based on the average height of all the skeletons in the training set:
        \begin{equation}
            (x_i, y_i)_j = \frac{(x_i, y_i)_j - (\overline{x}, \overline{y})_{pelvis}}{\overline{H}}
        \end{equation}
    where $(\overline{x}, \overline{y})_{pelvis}$ are the coordinates of the average pelvis in the training set and $\overline{H}$ is the average height of the skeletons in the training set;
    \item \textit{Global Coordinate Standardization} - Subtracting the mean and dividing by the standard deviation for each coordinate of the skeleton based on the statistics of the training set:
        \begin{equation}
            (x_i, y_i)_j = \frac{(x_i, y_i)_j - (\overline{x_i}, \overline{y_i})}{(\sigma(x_i), \sigma(y_i))}
        \end{equation}
    where $(\overline{x_i}, \overline{y_i})$ and $(\sigma(x_i), \sigma(y_i))$ are the average coordinates and their standard deviation computed for the $i$-th joint over the entire training set.

\end{itemize}

\textbf{Normalizations that remove position in the screen:}
\begin{itemize}
    \item \textit{Skeleton Level Translate} - Translating the $(x, y)$ coordinates of each skeleton with a different value - the $(x, y)$ coordinates of the pelvis for each individual skeleton:
        \begin{equation}
            (x_i, y_i)_j = (x_i, y_i)_j - (x_{pelvis}, y_{pelvis})_j
        \end{equation}
    \item \textit{Sequence Level Translate} - Translating the $(x, y)$ coordinates of each skeleton in the sequence with the same value - the $(x, y)$ coordinates of the pelvis of the middle skeleton in the sequence. This procedure removes the position information but preserves the direction of walking. This can be formulated as:
        \begin{equation}
            (x_i, y_i)_j = (x_i, y_i)_j - (x_{pelvis}, y_{pelvis})_{\frac{N}{2}}
        \end{equation}
\end{itemize}

\textbf{Normalizations that remove height information:}
\begin{itemize}

    \item \textit{Skeleton Level Scale} - Scaling the $(x, y)$ coordinates of each skeleton with a different value - the height of each corresponding skeleton in the sequence:
        \begin{equation}
            (x_i, y_i)_j = \frac{(x_i, y_i)_j}{H_j}
        \end{equation}
    where $H_j$ is the height of the current skeleton;
    \item \textit{Sequence Level Scale} - Scaling the $(x, y)$ coordinates of each skeleton in the sequence with the same value - the height of the middle skeleton in the sequence. This procedure removes the height information but preserves the variation of height in the sequence. It can be formulated as:
        \begin{equation}
            (x_i, y_i)_j = \frac{(x_i, y_i)_j}{ H_{\frac{N}{2}}}
        \end{equation}
    where $H_{\frac{N}{2}}$ is the height of the median skeleton in the sequence. 
\end{itemize}

\begin{figure}[hbt!]
    \centering
    \includegraphics[width=0.8\linewidth]{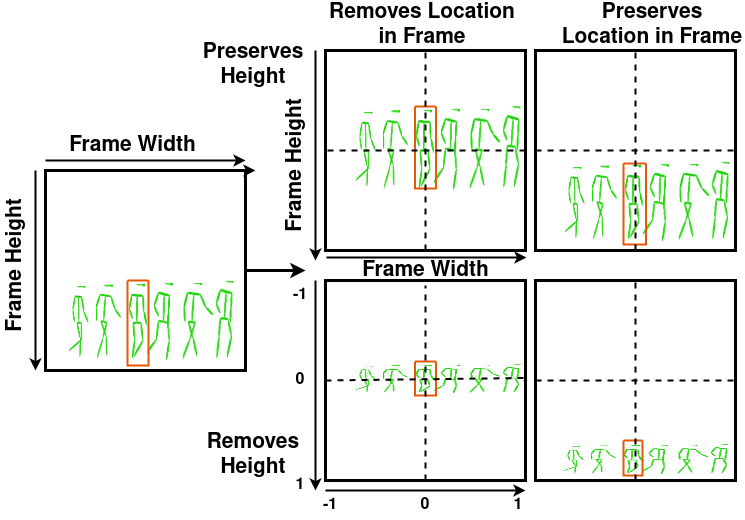}
    \caption{Illustrative example of the effects of different normalization schemes on a skeleton gait sequence. Translations remove the position on screen, while scaling removes some height information. Our experiments show that both procedures have a definite impact on downstream gait performance.}
    \label{fig:norm-examples}
\end{figure}

Figure \ref{fig:norm-examples} showcases an illustrative example on how scaling and translation affects the gait sequence. Scalings that remove height information distort the skeleton sequence making it harder to rely on height differences between individuals. Translations remove the position in the screen, which should intuitively not affect recognition. However, our results show the opposite, indicating the presence of spurious correlations in gait benchmarks.

We study the effects of the presented normalization techniques on the performance of the skeleton-based gait recognition models. We select 3 distinct popular skeleton-based gait recognition architectures: \textbf{GaitGraph} \cite{teepe2021gaitgraph} - a model that uses a spatio-temporal graph convolutions for learning discriminative gait features, \textbf{GaitFormer} \cite{cosma2022learning} - an architecture that learns temporal changes in consecutive skeletons with a transformer encoder disregarding explicit spatial relationships between joints, and \textbf{GaitPT} \cite{catruna2023gaitpt} - a hierarchical transformer that iteratively combines intermediary spatio-temporal features based on anatomical proximity.

We perform our analysis on 2 popular gait recognition benchmarks: \textbf{CASIA-B} \cite{CASIA:Yu}, a dataset constructed in controlled settings where subjects are recorded while walking from multiple angles with synchronized cameras and \textbf{GREW} \cite{GREW:zhu}, a dataset with people walking in the wild obtained from a wide range of surveillance cameras.


\subsection{Single Pose Model}
\begin{figure*}
\begin{center}
    \includegraphics[width=0.75\textwidth]{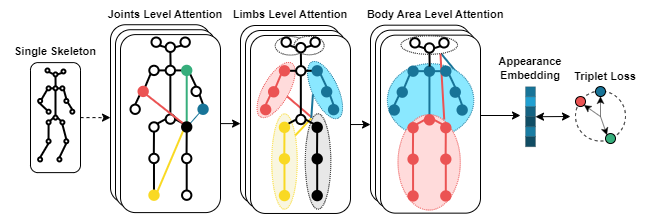}    
\end{center}
  \caption{The proposed architecture for extracting discriminative appearance information from a single pose. Our model computes self-attention at multiple levels to obtain the appearance representation of the skeleton.}
  \label{fig:no_gait_architecture}
\end{figure*}

We study how much discriminative information is encoded in a single skeleton. For this purpose we construct an architecture capable of identifying individuals by utilizing a single pose as input. As the model does not receive any type of temporal information, it cannot analyse the gait, and will instead focus only on spatial information and implicit appearance features. We construct a hierarchical transformer architecture for extracting the appearance information from a single skeleton.

The high-level overview of our proposed architecture is shown in Figure \ref{fig:no_gait_architecture}. The model initially computes multi-head self-attention between the joints of the input pose. The joint-level features are fused into groups of joints forming limb-level features based on anatomical proximity. At the second stage, multi-head self-attention is computed between the limb-level features. The resulting features are combined into limb group-level features, which are given as input to the final attention-based encoder that obtains the embedding. The embedding is obtained through the fusion of the final features into a single one and a linear projection.

Having as input a single 2D pose $P \in R^{18 \times 2}$, where $P = \{(x_1, y_1), (x_2, y_2),...(x_{18}, y_{18})\}$, our model projects the joint coordinates into a higher dimension and computes multi-head self-attention, treating each joint vector as an element in the sequence. This is formulated as:
\begin{equation}
    P' = MSA(PW^{C_1})
\end{equation}

where $MSA$ stands Multi-Head Self-Attention and $W^{C_1} \in R^{2 \times C_1}$ is the linear projection. The output feature map $P'=\{\boldsymbol{v_1}, \boldsymbol{v_2},...,\boldsymbol{v_{18}}\}$, where $ P' \in R^{18 \times C_1}$ goes through a merging module that combines vectors corresponding to neighboring joints. The output of the merging module can be written as:
\begin{equation}
    F_1 = \{(\boldsymbol{v_1} \mathbin\Vert\boldsymbol{v_2} \mathbin\Vert \boldsymbol{v_3})W^{C_2},..., (\boldsymbol{v_{16}} \mathbin\Vert \boldsymbol{v_{17}} \mathbin\Vert \boldsymbol{v_{18}})W^{C_2}\}
\end{equation}
where $\mathbin\Vert$ represents concatenation and each resulting feature vector in $F_1 \in R^{6 \times C_2}$ is obtained through the fusion of 3 joint-level feature vectors. The input vectors are arranged in such a manner that each 3 consecutive vectors correspond to the: left head, right head, left arm, right arm, left leg, right leg.

The same procedure is repeated with $F_1$ as input:
\begin{equation}
    F_1' = MSA(F_1)
\end{equation}
where $F_1' = \{\boldsymbol{v_1'}, \boldsymbol{v_2'},...,\boldsymbol{v_{6}'}\}$ and $F_1' \in R^{6\times C_2}$. These intermediary limb-level features are merged once again based on anatomical proximity with their corresponding limbs:
\begin{equation}
    F_2 = \{(\boldsymbol{v_1'} \mathbin\Vert \boldsymbol{v_2'})W^{C_3},..., (\boldsymbol{v_{5}'} \mathbin\Vert \boldsymbol{v_{5}'})W^{C_3}\}
\end{equation}
where the output is $F_2 \in R^{3 \times C_3}$ and $W^{C_3}$ is the linear projection. The resulting vectors correspond to: the head area, the upper body area, and the lower body area.

In the final stage, multi-head self-attention is computed on these area-level features:
\begin{equation}
    F_2' = MSA(F_2)
\end{equation}
with $F_2' = \{\boldsymbol{v_1''}, \boldsymbol{v_2''},\boldsymbol{v_{3}''}\}$ and $F_2' \in R^{3 \times C_3}$. The extracted features are merged into a single one through concatenation and are projected in to the embedding dimension $C_{emb}$:
\begin{equation}
    e = (\boldsymbol{v_1''} \mathbin\Vert \boldsymbol{v_2''} \mathbin\Vert \boldsymbol{v_3''})W^{emb}
\end{equation}
where the $e \in R^{emb}$ is a single high-dimensional vector that contains all discriminative appearance features. This embedding is given as input to the triplet loss for training the architecture.

The dimension of feature vectors starts from $C_1 = 32$ and gets doubled through every stage so that $C_2 = 64$ and $C_3 = 128$. The final embedding dimension after projection is also equal to $128$.
\subsection{Implementation Details}

We train GaitGraph, Gaitformer, GaitPT, and the single skeleton architecture by employing the triplet loss \cite{schroff2015facenet}, AdamW \cite{loshchilov2017decoupled} optimizer, and a cyclical learning rate policy \cite{smith2017cyclical}. In line with other works on gait recognition \cite{catruna2023gaitpt, teepe2021gaitgraph}, we chose a length of the gait sequence equal to 60 for CASIA-B and for GREW we opted for a sequence length equal to 30. For videos with less poses than the chosen sequence length, we interpolate between skeletons to obtain a sufficiently large sequence. In the case of videos with more poses, we randomly crop sequences for training and utilize the middle sequence for evaluation. The gait recognition architectures are trained for 300 epochs, while the single skeleton model for 30 epochs as the diversity of the training data is much lower resulting in earlier overfitting. We utilize minimal augmentation techniques in the form of temporal random cropping of walking sequences for the gait models and gaussian joint noise for all architectures. We train all models with 3 random seeds. All models have as output a normalized embedding that, after training, encodes the most discriminative features of the walking sequence. For recognition we follow the same approach as other methods \cite{catruna2023gaitpt,teepe2021gaitgraph} for gait identification: the identity prediction of a probe sample is done based on the nearest known neighbor in the embedding space, where known samples are those in the gallery set.  

Note that our results might differ from reported results for each architecture since we train with a slightly different set of hyperparameters and data augmentation schemes compared to the original authors. However, our analysis focuses on a fair comparison between the overall effects of normalization procedures and not between architectures.

\section{Results}
\label{sec:results}

For evaluating the normalization techniques on the CASIA-B dataset, we follow the same evaluation protocol as GaitGraph \cite{teepe2021gaitgraph} and GaitPT \cite{catruna2023gaitpt}. On the testing set, pairs of probe and galleries are constructed based on all possible combinations of angles and walking scenarios (NM, CL and BG), excluding identical view cases. The accuracy is computed individually for each (probe, gallery) pair and the performance for a certain angle can be calculated as the average of accuracies of all the pairs containing that probe view. Similarly, the accuracy for a certain walking scenario can be computed as the average of the accuracies of all probe pairs with that specific setting. For brevity and comparison purposes, we only show the average recognition accuracy across all gallery-probe pairs for each model and normalization technique. As laid out in Section \ref{sec:method}, we divide the normalization techniques into 4 groups: those which remove height information by scaling, those which remove position information by translation, combinations of scaling and translation which remove both height and position information, and those which preserve all information by scaling and translating with constant values. 

\begin{table*}[hbt!]
    \caption{Comparison of the experimental normalization techniques on the CASIA-B test set. We report the average recognition accuracy across all probe scenarios and angles excluding the identical views. For all runs we compute the mean and standard deviation of 3 runs with different random seeds.}
    \label{tab:normalizations_casia}
    \begin{center}
    \resizebox{0.95\textwidth}{!}{
    \begin{tabular}{l | l |  c | c | c | c}
         \textbf{Effect} & \textbf{Normalization} & \textbf{GaitGraph} &  \textbf{GaitFormer} & \textbf{GaitPT} & \textbf{Mean} \\
        \midrule
        \multirow{2}{*}{Removes Height } & \textit{Sequence Level Scale} & 70.65 \scriptsize{$\pm$ 1.07} & 50.19 \scriptsize{$\pm$ 2.64} & 74.82 \scriptsize{$\pm$ 2.81} & 65.22 \scriptsize{$\pm$ 2.17} \\
        & \textbf{\large \color{red} !} \textit{Skeleton Level Scale}  & 61.79 \scriptsize{$\pm$ 2.37} & 45.19 \scriptsize{$\pm$ 2.44} & 58.44 \scriptsize{$\pm$ 2.64} & \underline{55.14} \scriptsize{$\pm$ 2.48} \\
        \midrule
        \multirow{2}{*}{Removes Position} & \textit{Sequence Level Translate} & 75.84 \scriptsize{$\pm$ 0.49} & 53.84 \scriptsize{$\pm$ 1.68} & 64.75 \scriptsize{$\pm$ 1.9} & 64.81 \scriptsize{$\pm$ 1.36} \\
        & \textbf{\large \color{red} !} \textit{Skeleton Level Translate} & 61.94 \scriptsize{$\pm$ 0.31} & 33.44 \scriptsize{$\pm$ 0.55} & 54.87 \scriptsize{$\pm$ 2.55} & \underline{50.09} \scriptsize{$\pm$ 1.14} \\
        \midrule
        \multirow{4}{*}{Removes Position + Height} & \textit{Sequence Level Translate + Sequence Level Scale} & 75.85 \scriptsize{$\pm$ 0.7} & 51.22 \scriptsize{$\pm$ 0.99} & 72.04 \scriptsize{$\pm$ 3.21} & 66.37 \scriptsize{$\pm$ 1.63} \\
        & \textbf{\large \color{red} !} \textit{Skeleton Level Translate + Skeleton Level Scale} & 59.83 \scriptsize{$\pm$ 0.7} & 21.71 \scriptsize{$\pm$ 0.74} & 48.84 \scriptsize{$\pm$ 2.53} & \underline{43.46} \scriptsize{$\pm$ 1.32} \\
        & \textbf{\large \color{red} !} \textit{Skeleton Level Translate + Sequence Level Scale} & 60.68 \scriptsize{$\pm$ 1.02} & 22.52 \scriptsize{$\pm$ 0.63} & 46.12 \scriptsize{$\pm$ 1.53} & \underline{43.11} \scriptsize{$\pm$ 1.06} \\
        & \textbf{\large \color{red} !} \textit{Sequence Level Translate + Skeleton Level Scale} & 66.63 \scriptsize{$\pm$ 1.32} & 43.89 \scriptsize{$\pm$ 0.59} & 58.78 \scriptsize{$\pm$ 1.38} & \underline{56.43} \scriptsize{$\pm$ 1.10} \\
        \midrule
        \multirow{4}{*}{Preserves All Info} & \textit{Frame Scale} & 68.22 \scriptsize{$\pm$ 0.65} & 55.09 \scriptsize{$\pm$ 2.96} & 77.96 \scriptsize{$\pm$ 4.3} & 67.09 \scriptsize{$\pm$ 2.63} \\
        & \textit{BatchNorm} & 65.95 \scriptsize{$\pm$ 1.57} & 59.53 \scriptsize{$\pm$ 0.3} & 80.28 \scriptsize{$\pm$ 0.88} & 68.59 \scriptsize{$\pm$ 0.92} \\
        & \textit{Global Average Skeleton} & 68.04 \scriptsize{$\pm$ 1.27} & 55.32 \scriptsize{$\pm$ 0.7} & 81.51 \scriptsize{$\pm$ 0.39} & 68.29 \scriptsize{$\pm$ 0.79} \\
        & \textit{Global Coordinate Standardization} & 65.34 \scriptsize{$\pm$ 1.26} & 61.77 \scriptsize{$\pm$ 0.88} & 79.0 \scriptsize{$\pm$ 0.81} & 68.71 \scriptsize{$\pm$ 0.98} \\
    \end{tabular}
    }
    \end{center}
\end{table*}

\noindent \textbf{Removing height and position in frame severely affects performance in controlled scenarios.}
Table \ref{tab:normalizations_casia} shows the performance of each normalization approach for every experimental skeleton gait recognition architecture. For a fair comparison between normalization techniques, we utilized the same training loss, sequence length, augmentations and hyperparameters which might differ to those employed in the official implementations \cite{teepe2021gaitgraph,cosma2022learning,catruna2023gaitpt}. 

For height removal normalization techniques, Sequence Level Scaling has close performance to those which preserve all information. However, Skeleton Level Scaling obtains a much lower recognition accuracy, showing how eliminating the height information and height variation can lead to a loss in performance. 

In terms of position removal normalizations, Skeleton Level Translation also obtains a decrease in accuracy when compared to Sequence Level Translation. Two possible explanations are: (1) the model loses the location information based on which it can make correlations across different views, or (2) the model loses information about the direction in which a person is walking, which can be important when analysing the gait. We consider the second explanation weaker as the walking direction can be inferred based on the orientation of individual skeletons. In terms of combinations of translation and scaling, it seems that any type of scaling with the Skeleton Level Translation leads to worse performance. This indicates that the gait recognition models learn to make predictions by associating certain positions across all the cameras. 

\noindent \textbf{Preserving height and position in frame boosts performance.}
The final normalization techniques (i.e., Frame Scale, BatchNorm, Global Average Skeleton and Global Coordinate Standardization) seem to obtain a fairly equal performance, which is to be expected as all four share an important aspect: they preserve all height and position information by scaling and translating with constant values. The only difference is how these constant values are computed: Frame Scale involves constants computed based on screen size, for BatchNorm the model learns the scaling and translation during training, for Global Average Skeleton the values are computed based on the average pelvis and average height of all poses in the training set, and for Global Coordinate Standardization the values are computed based on all the individual joints in the training set. The fact that these four approaches use constant values for normalization means that the model does not lose any height and location information. Consequently, these approaches consistently obtain the highest recognition accuracy. 

These results strongly point towards the presence of spurious correlations arising from the construction of CASIA-B, most likely due to the use of synchronized cameras, which introduced unintended cues that aid recognition.


\begin{table*}[hbt!]
    \caption{Comparison of the experimental normalization techniques on the GREW test set. We report the mean and standard deviation of the rank-1 average recognition accuracy computed over 3 different random seeds.}
    \label{tab:normalizations_grew}
    \begin{center}
    \resizebox{0.95\textwidth}{!}{
    \begin{tabular}{l | l |  c | c | c | c}
         \textbf{Effect} & \textbf{Normalization} & \textbf{GaitGraph} &  \textbf{GaitFormer} & \textbf{GaitPT} & \textbf{Mean R1 Accuracy} \\
        \midrule
        \multirow{2}{*}{Removes Height} & \textit{Sequence Level Scale} & 20.31 \scriptsize{$\pm$ 0.39} & 7.65 \scriptsize{$\pm$ 0.35} & 31.87 \scriptsize{$\pm$ 0.58} & 19.94 \scriptsize{$\pm$ 0.44} \\
        & \textit{Skeleton Level Scale} & 20.59 \scriptsize{$\pm$ 0.8} & 8.19 \scriptsize{$\pm$ 0.42} & 31.65 \scriptsize{$\pm$ 0.06} & 20.14 \scriptsize{$\pm$ 0.43} \\
        \midrule
        \multirow{2}{*}{Removes Position} & \textit{Sequence Level Translate} & 17.65 \scriptsize{$\pm$ 0.19} & 17.18 \scriptsize{$\pm$ 0.29} & 28.56 \scriptsize{$\pm$ 0.21} & 21.13 \scriptsize{$\pm$ 0.23} \\
        & \textit{Skeleton Level Translate} & 17.25 \scriptsize{$\pm$ 0.35} & 18.4 \scriptsize{$\pm$ 0.42} & 27.5 \scriptsize{$\pm$ 1.51} & 21.05 \scriptsize{$\pm$ 0.76} \\
        \midrule
        \multirow{4}{*}{Removes Position + Height} &  \textit{Sequence Level Translate + Sequence Level Scale} & 20.6 \scriptsize{$\pm$ 0.26} & 7.87 \scriptsize{$\pm$ 0.23} & 31.13 \scriptsize{$\pm$ 0.67} & \underline{19.87} \scriptsize{$\pm$ 0.39} \\
        &  \textit{Skeleton Level Translate + Skeleton Level Scale} & 20.28 \scriptsize{$\pm$ 0.34} & 8.02 \scriptsize{$\pm$ 0.5} & 31.18 \scriptsize{$\pm$ 0.36} & \underline{19.83} \scriptsize{$\pm$ 0.4} \\
        &    \textit{Skeleton Level Translate + Sequence Level Scale} & 20.06 \scriptsize{$\pm$ 0.55} & 7.49 \scriptsize{$\pm$ 0.74} & 31.66 \scriptsize{$\pm$ 0.5} & \underline{19.74} \scriptsize{$\pm$ 0.6} \\
        &  \textit{Sequence Level Translate + Skeleton Level Scale} & 20.57 \scriptsize{$\pm$ 0.55} & 8.05 \scriptsize{$\pm$ 0.32} & 30.44 \scriptsize{$\pm$ 0.43} & \underline{19.68} \scriptsize{$\pm$ 0.43} \\
        \midrule
        \multirow{4}{*}{Preserves All Info} & \textit{BatchNorm} & 21.07 \scriptsize{$\pm$ 0.56} & 16.6 \scriptsize{$\pm$ 0.91} & 34.15 \scriptsize{$\pm$ 0.48} & 23.94 \scriptsize{$\pm$ 0.65} \\
        & \textit{Global Average Skeleton} & 20.32 \scriptsize{$\pm$ 0.98} & 7.55 \scriptsize{$\pm$ 0.31} & 32.57 \scriptsize{$\pm$ 0.36} & 20.15 \scriptsize{$\pm$ 0.55} \\
        & \textit{Global Coordinate Standardization} & 19.94 \scriptsize{$\pm$ 0.26} & 14.54 \scriptsize{$\pm$ 0.5} & 34.68 \scriptsize{$\pm$ 0.79} & 23.05 \scriptsize{$\pm$ 0.52} \\
    \end{tabular}
    }
    \end{center}
\end{table*}

\noindent \textbf{In-the-wild datasets force the model to process temporal dynamics.}
Table \ref{tab:normalizations_grew} shows the performance of the analysed normalization techniques on the GREW dataset. As the testing probe identities are not publicly released, for our analysis, we construct a testing set by dividing the gallery set into both gallery and probe. For each unique identity, half of the walking sequences are placed into gallery and the remaining half into the probe set. Identities which have only one sequence remain only in the gallery. Consequently, the performance of the experimental architecture differ from their official results as the results are reported on a different test set. As there is no information about the size of the frames of the videos, we omit the Frame Scale normalization. 

In contrast to the results on CASIA-B, the normalization techniques achieve similar performance on the GREW test set. Each of the experimental architectures seems to have a different setting in which it performs the worst: For GaitGraph and GaitPT the worst settings are the normalizations which only remove position information by scaling, while for GaitFormer the lowest performance involves all normalizations which include removing height information by scaling. The top-performing normalization techniques overall are the Batch Normalization and standardization based on Average Coordinates. The difference between their accuracy and the accuracy of the worst-performing normalization is approximately 4\%. These results indicate that, for more difficult, in-the-wild data, the models cannot rely only on appearance-based features and may be forced to process temporal dynamics of gait as well. This is a possible explanation for the significantly lower performance of pose-based methods compared to silhouette-based architectures \cite{fan2020gaitpart,lin2022gaitgl} on these datasets.

\begin{figure}[hbt!]
    \centering
    \includegraphics[width=0.85\linewidth]{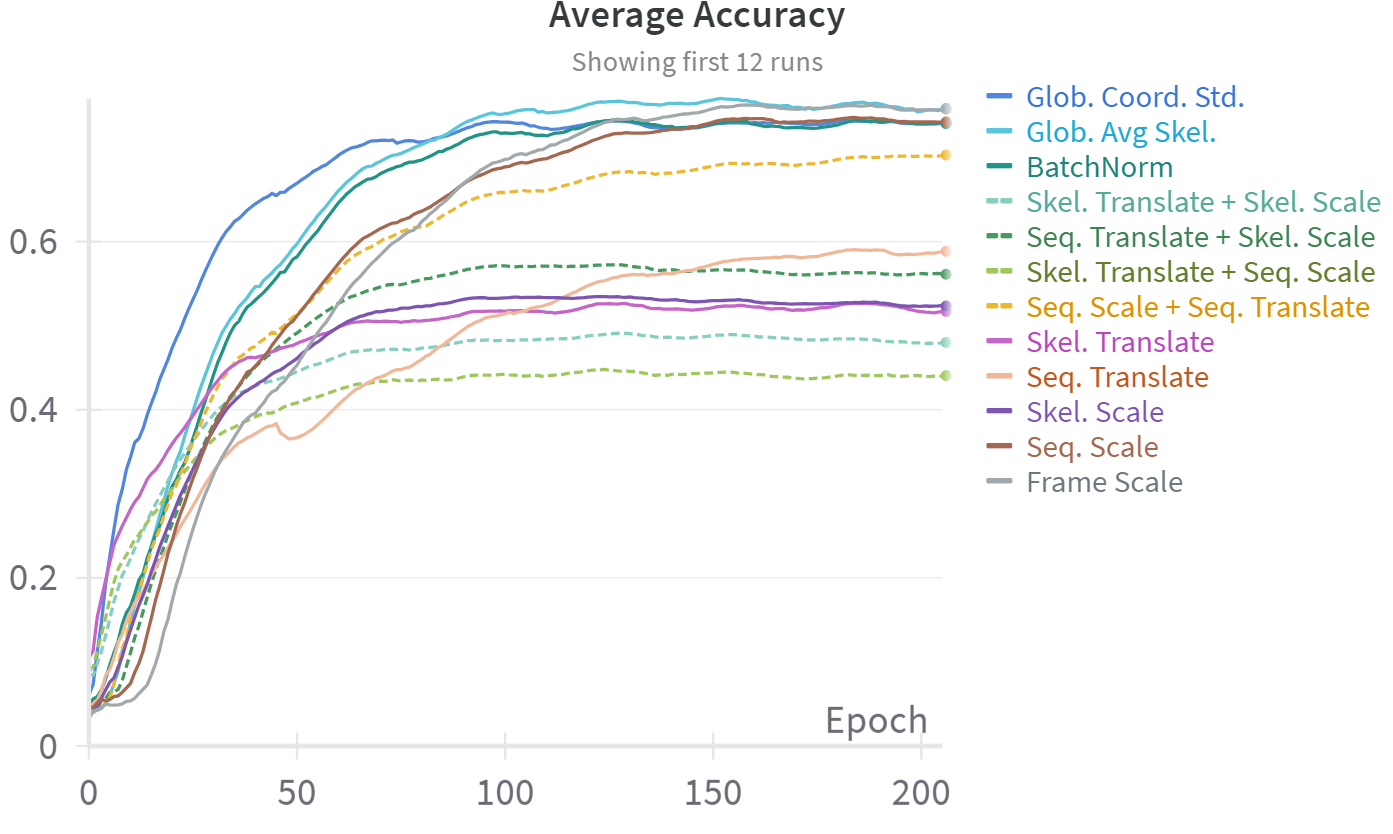}
    \caption{Average testing accuracy for all normalization techniques during training on CASIA-B.}
    \label{fig:gaitpt-performance-casia}
\end{figure}

\noindent \textbf{Quantity of preserved appearance information improves recognition performance during training.}
Figure \ref{fig:gaitpt-performance-casia} shows the average recognition accuracy of the GaitPT architecture during its training for each normalization setting on CASIA-B. From the figure, we can divide the normalizations in 2 groups based on performance: normalizations that preserve all position and height-related information and obtain the highest performance, and normalizations that remove relevant appearance and location information and obtain a lower performance. From the figure we can deduce that the quantity of removed information regarding appearance and position is correlated to a worse performance. This can be observed in normalizations that remove only some information such as Sequence Level Scale + Sequence Level Translate, which obtains a higher accuracy compared to the worst-performing normalizations, but a lower performance compared to the top-performing ones. 

\begin{figure}[hbt!]
    \centering
    \includegraphics[width=0.85\linewidth]{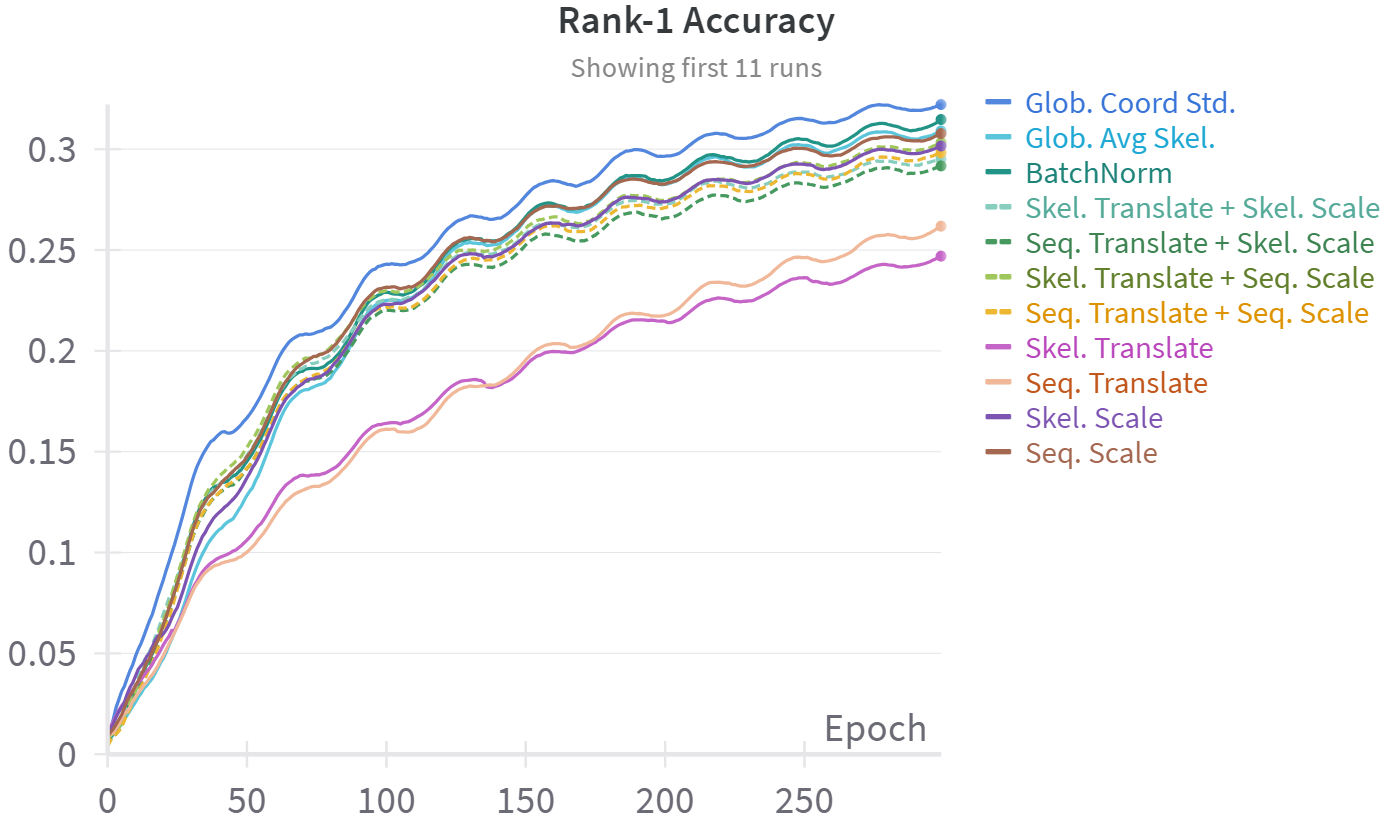}
    \caption{Rank-1 accuracy for all normalization techniques during training on GREW.}
    \label{fig:gaitpt-performance-grew}
\end{figure}

We show the same performance over training epochs on the GREW dataset in Figure \ref{fig:gaitpt-performance-grew}. In this case, there is less discrepancy between normalization approaches. The accuracy does not seem correlated to the quantity of appearance and position information, indicating that, for in-the-wild datasets, such information has less impact in identification. This shows that gait recognition models find shortcuts in recognizing people on data obtained in controlled laboratory-settings such as CASIA-B. 

\noindent \textbf{A single pose reveals the identity unreasonably well in both controlled and in-the-wild scenarios.}
We follow the same evaluation protocol on CASIA-B for the single skeleton model. We analyse the performance of two testing approaches. In the first approach, the model outputs the embedding prediction based on the middle pose in the walking sequence. In the second approach, the model computes a prediction for each pose in the sequence and the final embedding is an average over all the poses in the sample. We call this second approach Average Prediction. Both of these approaches also use \textit{no temporal information} as each pose is processed independently by the model. We experiment with the same types of normalization techniques on the single skeleton model as well. As the input to the model is only one pose, the sequence level normalization approaches are the same as the skeleton level ones, therefore omitted.

\begin{table*}[hbt!]
    \caption{Results of the single skeleton model for all types of normalization techniques on the CASIA-B test set. The best model manages to obtain a recognition accuracy of 57.49\% using only one pose, demonstrating that a single skeleton encodes discriminative appearance information.}
    \label{tab:no_gait_casia}
    \begin{center}
    \resizebox{\textwidth}{!}{
    \begin{tabular}{ l | c | c | c | c | c | c | c | c }
        & \multicolumn{4}{c|}{\textbf{Single Skeleton}} & \multicolumn{4}{c}{\textbf{Average Prediction over Multiple Single Skeletons}} \\
         \textbf{Normalization} & \textbf{NM} &  \textbf{BG} & \textbf{CL} & \textbf{Mean} & \textbf{NM} & \textbf{BG} & \textbf{CL} & \textbf{Mean} \\
        \midrule
    \textit{Skeleton Level Scale} & 27.43 \scriptsize{$\pm$ 0.6} & 21.7 \scriptsize{$\pm$ 0.85} & 14.96 \scriptsize{$\pm$ 0.91} & 21.37 \scriptsize{$\pm$ 0.57} & 64.81 \scriptsize{$\pm$ 0.86} & 48.24 \scriptsize{$\pm$ 1.65} & 32.95 \scriptsize{$\pm$ 1.63} & 48.66 \scriptsize{$\pm$ 0.59} \\
    \midrule
    \textit{Skeleton Level Translate} & 18.35 \scriptsize{$\pm$ 1.08} & 12.55 \scriptsize{$\pm$ 0.71} & 8.72 \scriptsize{$\pm$ 0.69} & 13.21 \scriptsize{$\pm$ 0.41} & 59.53 \scriptsize{$\pm$ 1.15} & 41.19 \scriptsize{$\pm$ 0.68} & 25.05 \scriptsize{$\pm$ 0.14} & 41.93 \scriptsize{$\pm$ 0.54} \\
    \midrule
    \textit{Skel. Lvl. Translate + Skel. Lvl. Scale} & 19.21 \scriptsize{$\pm$ 0.53} & 13.71 \scriptsize{$\pm$ 0.4} & 8.07 \scriptsize{$\pm$ 0.14} & 13.66 \scriptsize{$\pm$ 0.17} & 60.04 \scriptsize{$\pm$ 0.54} & 39.87 \scriptsize{$\pm$ 0.76} & 21.06 \scriptsize{$\pm$ 0.97} & 40.33 \scriptsize{$\pm$ 0.29} \\
    \textit{Skel. Lvl Translate + Frame Scale} & 19.93 \scriptsize{$\pm$ 0.28} & 14.68 \scriptsize{$\pm$ 0.67} & 8.95 \scriptsize{$\pm$ 0.32} & 14.5 \scriptsize{$\pm$ 0.36} & 60.07 \scriptsize{$\pm$ 0.84} & 40.76 \scriptsize{$\pm$ 0.84} & 23.4 \scriptsize{$\pm$ 0.47} & 41.41 \scriptsize{$\pm$ 0.12} \\
    \midrule
    \textit{Frame Scale} & \textbf{36.45} \scriptsize{$\pm$ 2.67} & \textbf{30.99} \scriptsize{$\pm$ 1.6} & \textbf{23.28} \scriptsize{$\pm$ 0.41} & \textbf{30.24} \scriptsize{$\pm$ 1.56} & 70.71 \scriptsize{$\pm$ 1.13} & 56.68 \scriptsize{$\pm$ 1.19} & 41.44 \scriptsize{$\pm$ 0.77} & 56.28 \scriptsize{$\pm$ 0.87} \\
    \textit{BatchNorm} & 36.34 \scriptsize{$\pm$ 0.65} & 30.12 \scriptsize{$\pm$ 0.6} & 21.73 \scriptsize{$\pm$ 0.85} & 29.4 \scriptsize{$\pm$ 0.2} & \textbf{73.0} \scriptsize{$\pm$ 0.58} & \textbf{57.95} \scriptsize{$\pm$ 0.18} & \textbf{41.53} \scriptsize{$\pm$ 0.22} & \textbf{57.49} \scriptsize{$\pm$ 0.33} \\
    \textit{Global Average Skeleton} & 36.08 \scriptsize{$\pm$ 0.88} & 30.2 \scriptsize{$\pm$ 0.73} & 23.23 \scriptsize{$\pm$ 1.06} & 29.66 \scriptsize{$\pm$ 0.55} & 71.2 \scriptsize{$\pm$ 0.69} & 56.92 \scriptsize{$\pm$ 1.09} & 40.12 \scriptsize{$\pm$ 0.6} & 56.08 \scriptsize{$\pm$ 0.72} \\
    \textit{Global Coordinate Standardization} & 33.55 \scriptsize{$\pm$ 1.32} & 28.49 \scriptsize{$\pm$ 0.75} & 21.33 \scriptsize{$\pm$ 1.21} & 27.79 \scriptsize{$\pm$ 1.0} & 71.41 \scriptsize{$\pm$ 1.32} & 56.11 \scriptsize{$\pm$ 0.86} & 41.58 \scriptsize{$\pm$ 0.78} & 56.37 \scriptsize{$\pm$ 0.66} \\
    \end{tabular}
    }
    \end{center}
\end{table*}

Table \ref{tab:no_gait_casia} presents the performance of the single skeleton model for all types of normalization techniques on the scenarios of the CASIA-B test set following the exact evaluation protocol as for gait recognition. For single skeleton input, the top performance of 30.24\% accuracy is obtained with the Frame Scale normalization. However, the BatchNorm, Global Average Skeleton, and Global Coordinate Standardization obtain a comparable performance and the slight difference in accuracy may be due to favourable random seeds. In contrast, the normalization that removes the most appearance information (i.e. Skeleton Level Translate + Skeleton Level Scale) obtains the lowest accuracy of 13.21\%. It is worth mentioning that this accuracy is much better than random, meaning that some appearance information (e.g. limb proportions) is still encoded in the samples. In the case of Average Prediction, the order of performance remains similar but the accuracy increases to a top-performance of 57.49\%. For reference, the state-of-the-art architecture in skeleton-based gait recognition obtains an accuracy of approximately 82.5\% \cite{catruna2023gaitpt}. This indicates that gait information accounts for approximately 30\% of the extracted discriminative features on CASIA-B.


\begin{table}[hbt!]
    \caption{Results of the single skeleton model for all types of normalization techniques on the GREW test set.}
    \label{tab:no_gait_grew}
    \begin{center}
    \resizebox{0.90\linewidth}{!}{
    \begin{tabular}{ p{3.4cm} | c | c | c | c }
        & \multicolumn{2}{c|}{\textbf{Single Skeleton}} & \multicolumn{2}{p{3cm}}{\textbf{Average Prediction over Multiple Single Skeletons}} \\
         \textbf{Normalization} & \textbf{R1} &  \textbf{R5} & \textbf{R1} & \textbf{R5} \\
        \midrule
        \textit{Skeleton Level Scale} & 7.3 \scriptsize{$\pm$ 0.58} & 16.07 \scriptsize{$\pm$ 1.06} & 24.32 \scriptsize{$\pm$ 1.41} & 40.35 \scriptsize{$\pm$ 1.91} \\
        \midrule
    \textit{Skeleton Level Translate} & 7.14 \scriptsize{$\pm$ 0.19} & 16.48 \scriptsize{$\pm$ 0.69} & 23.45 \scriptsize{$\pm$ 0.16} & 39.64 \scriptsize{$\pm$ 0.28} \\
    \midrule
    \textit{Skel. Lvl. Translate + Skel. Lvl Scale} & 6.94 \scriptsize{$\pm$ 0.14} & 15.72 \scriptsize{$\pm$ 0.09} & 23.99 \scriptsize{$\pm$ 0.07} & 39.43 \scriptsize{$\pm$ 0.15} \\
    \midrule
    \textit{BatchNorm} & 8.0 \scriptsize{$\pm$ 0.09} & 17.63 \scriptsize{$\pm$ 0.31} & \textbf{25.3} \scriptsize{$\pm$ 0.19} & 41.7 \scriptsize{$\pm$ 0.33} \\
    \textit{Global Average Skeleton} & 8.89 \scriptsize{$\pm$ 0.31} & 19.76 \scriptsize{$\pm$ 0.4} & \textbf{28.19} \scriptsize{$\pm$ 0.25} & 45.15 \scriptsize{$\pm$ 0.44} \\
    \textit{Glob. Coord. Std.} & 8.27 \scriptsize{$\pm$ 0.52} & 18.28 \scriptsize{$\pm$ 1.15} & \textbf{26.29} \scriptsize{$\pm$ 1.88} & 43.23 \scriptsize{$\pm$ 2.11} \\
    \end{tabular}
    }
    \end{center}
\end{table}

As the single pose model obtains unreasonably good recognition performance on CASIA-B, we evaluate it on the GREW dataset following the same evaluation protocol as for the normalization techniques. Table \ref{tab:no_gait_grew} shows the results of the model in terms of R-1 and R-5 accuracy for both an input of a single skeleton and Average Prediction.

The results show that the single skeleton model is able to use the appearance information encoded in individual poses to identify people in the wild as well. By analysing a single skeleton, the model obtains a 8.89\% Rank-1 accuracy and 19.76\% Rank-5 accuracy with the top-performing normalization of translating based on the global average pelvis coordinate and scaling based on the global average skeleton height. In the case of Average Prediction where all poses in the sequence are analysed individually and their embeddings are averaged, the model manages to obtain a Rank-1 accuracy of 28.19\% surpassing the GaitGraph and GaitFormer architectures. This is a surprising result as the single skeleton model does not take into account temporal information. 
Normalization techniques have little impact as the worst-performing and top-performing approaches for the single skeleton model have similar performance. An explaination is that the model primarily focuses on other appearance features such as limb proportions, which are retained through all experimental normalizations.

\section{Discussion and Conclusions}
\label{sec:conclusions}
It remains unclear whether the improvements in skeleton gait recognition are attributed to motion patterns or if there is an information leakage of anthropometric information from modality extractors. Our experiments show that removing height information and position on the screen markedly lowers performance in gait recognition. Moreover, in some scenarios, we showed that recognition is feasible using only a single pose, disregarding temporal dynamics of the walk. 

The fundamental challenge of gait recognition, the disentanglement of appearance factors from movement, appears to remain unsolved. Furthermore, current datasets captured in controlled settings  (e.g., CASIA-B \cite{CASIA:Yu}) appear to provide "shortcuts", enabling the model to learn spurious correlations that reflect the data collection procedure: i.e. synchronized cameras, constant subject distance from the camera and others. Our experiments show that more diverse "in-the-wild" datasets such as GREW \cite{GREW:zhu} are less prone to shortcuts, indicating a pressing need for more such resources to be made available to the research community.

For gait recognition in practical scenarios, we consider that there is a trade-off between accuracy and the use of anthropometric data. Striking a balance between isolating motion information and utilizing appearance data is crucial for achieving robust gait analysis.



{\small
\bibliographystyle{ieee}
\bibliography{refs}
}

\end{document}